# Power efficient Spiking Neural Network Classifier based on memristive crossbar network for spike sorting application

Anand Kumar Mukhopadhyay[1], *Graduate Student Member*, *IEEE*, Indrajit Chakrabarti[2], *Member*, *IEEE*, Arindam Basu[3], *Member*, *IEEE*, and Mrigank Sharad[4]

***Abstract*—In this paper, authors have presented a power efficient scheme for implementing a spike sorting module. Spike sorting is an important application in the field of neural signal acquisition for implantable biomedical systems whose function is to map the Neural-spikes (N-spikes) correctly to the neurons from which it originates. The accurate classification is a pre-requisite for the succeeding systems needed in Brain-Machine-Interfaces (BMIs) to give better performance. The primary design constraint to be satisfied for the spike sorter module is low power with good accuracy. There lies a trade-off in terms of power consumption between the on-chip and off-chip training of the N-spike features. In the former case, care has to be taken to make the computational unit's power efficient whereas in the later the data rate of wireless transmission should be minimized to reduce the power consumption due to the transceivers. In this work, a 2-step shared training scheme involving a K-means sorter and a Spiking Neural Network (SNN) is elaborated for on-chip training and classification. Also, a low power SNN classifier scheme using memristive crossbar type architecture is compared with a fully digital implementation. The advantage of the former classifier is that it is power efficient while providing comparable accuracy as that of the digital implementation due to the robustness of the SNN training algorithm which has a good tolerance for variation in memristance.***

## I. Introduction

THE idea of implementing a neural network in hardware was started long back in the late 1990's by Carver Mead, who opened up the field of neuromorphic engineering. He showed that biological neuronal properties can be replicated to a greater extent by using large-scale adaptive analog systems as they dissipate less power and are more robust to component degradation and failure [1]. A multidisciplinary approach is essential for facing the challenges in the field of neuromorphic computing since there lies scope for developing better hardware with the advancement in technology in the field of Very Large Scale Integration (VLSI) [2].

To make a neural network-based system in hardware requires a considerable number of transistors as the processing elements. Compact implementation of these systems is possible because of greater integration density of the transistors using silicon-based nano-scaled Metal Oxide Semiconductor Field Effect Transistors (MOSFETs). However, when compared to the human brain in terms of integration density, the complex neuromorphic systems built using transistors replicating the behavior of neurons and synapses is far behind. Therefore, for making the implementation further compact hybrid techniques have been adopted by hetero-integration of Complementary Metal Oxide Semiconductor (CMOS) with emerging devices like memristor [3].

The concept of memristor was first proposed in 1971 by Leon O Chua, is widely being used in recent times in neuromorphic systems having an important function of representing the synaptic weights [4][5]. A memristor is a non-volatile device, also known as memory resistor whose resistance can be precisely modulated by passing current pulses through it. Since memristors are nanoscaled devices and also compatible with silicon, they have attracted researchers in the field of neuromorphic engineering to make use of it in systems which can function similarly like a human brain for classifying objects and recognizing patterns [6][7][8]. These memristor act as a bridge in a crossbar type architecture connecting the pre-synaptic neurons to the post-synaptic neurons making a compact implementation possible in hardware [9] [10].

Sung Hyun Jo et al. in 2008 have shown that a memory density of 2 Gbits/cm$^2$ is possible using Ag and poly-Si (p-Si) as the perpendicular nanowires with amorphous-Si (a-Si) acting as the memristive switches [11]. These crossbar arrays using nanowires is extremely useful in implementing low power, area-constrained application specific hardware architectures requiring a considerable amount of computation and storage of memory bits.

Spiking Neural Network (SNN), which is also the 3rd generation of neural network is more superior to the previous generations of neural network when compared with the

Authors 1, 2, 4 are with the department of Electronics and Electrical Communication Engineering , Indian Institute of Technology, Kharagpur, West Bengal, 721302 India (e-mail: anandmukh@iitkgp.ac.in; indrajit@ece.iitkgp.ac.in; mrigank@ece.iitkgp.ac.in).

Author 3 is with the department of Electrical and Electronic Engineering, Nanyang Technological University, 639798 Singapore (e-mail: arindam.basu@ntu.edu.sg).

biological brain [12] [13] [14]. The behavior of a spiking neuron has been represented by different mathematical models having different spiking natures and computational complexity [15] [16]. However, for practical computation, the significantly simpler models may suffice which capture the basic characteristics of leaky integrate and fire [17], clubbed with a spike-timing-based learning algorithm. It has been seen that using STDP and Hebbian learning as the training algorithms on a dataset represented by 1's and 0's, trained binary weights could be generated having a good amount of accuracy [18][19].

Spike sorting is an important application the field of neural signal acquisition wherein spikes present in neural signals are to be mapped correctly to the neurons from which it originates [23][28][29]. For realizing the spike sorting as an implantable device such that it can function in real-time acting as a Brain Machine Interface (BMI), energy efficiency with considerable good accuracy becomes the primary design constraint [24, 25].

Maryam Saeed *et al.* made a comparison of five different classifier architectures for online neural spike sorting based on accuracy and computational complexity in which their analysis was done while considering the training off-chip and using the trained parameters by the classifier on-chip for avoiding the power dissipation due to the computationally intensive training process [26]. But, it is to be noted that a significant amount of power is dissipated due to the wireless transmission of data by the transceivers [27]. For avoiding the power due to wireless data transfers, many researchers have considered a fully on-chip spike sorter, having not only the classifier but also the trainer within the chip which is indeed a challenging task while meeting the major design constraints of low power and high classification accuracy [24 30].

Recently, Rakshit *et al.* proposed a low power 2-step shared training scheme in which the unsupervised data is initially clustered by a K-means sorter block, after which a supervised power efficient SNN trainer computes the trained weights [19]. In this paper, authors have elaborated the system level spike sorting module proposed earlier. The major contributions of the paper are as follows:

1) The power consumption by the major blocks involved in the spike sorting module has been discussed.

2) An alternate low power scheme for representing the SNN classifier using a memristive crossbar network has been compared with the one proposed earlier [19].

In section II, an overview of the system level description of the Neural-spike sorting module is presented. The shared-2-step training module is discussed in section III. Then a simple mathematical model of a memristive crossbar network is discussed in Section IV. In section V, two different architectural schemes of the SNN classifier is elaborated. In section VI, the need for adaptation and retraining is mentioned. In section VII, results and discussions on the two types of SNN classifier: 1) fully digital, 2) memristive crossbar network are compared in terms of power and also the effect of variation of memristance on accuracy is studied. Finally, we conclude by summarizing important points in section VIII.

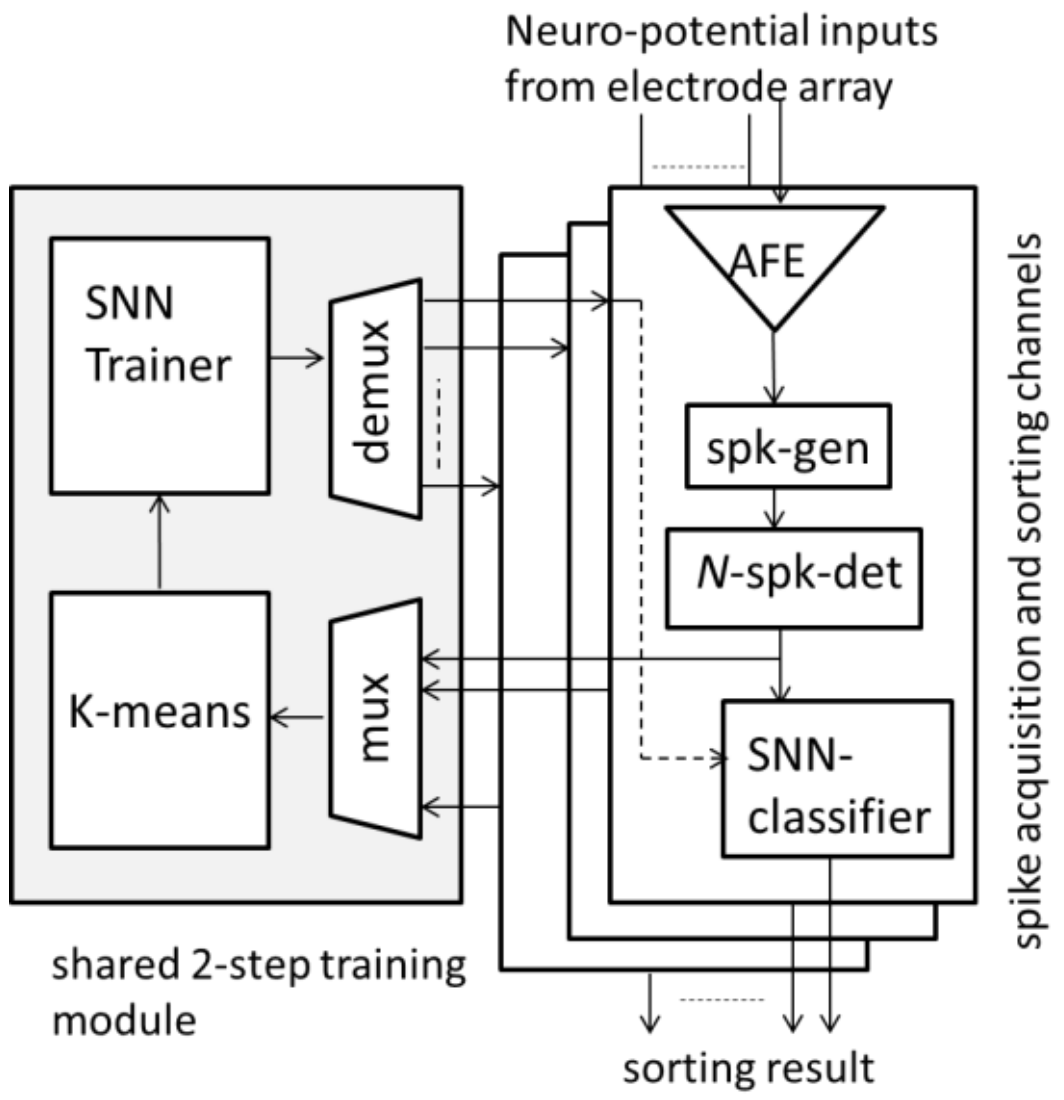

Fig. 1. System level diagram of proposed spike sorting scheme with 2-step training module [19].

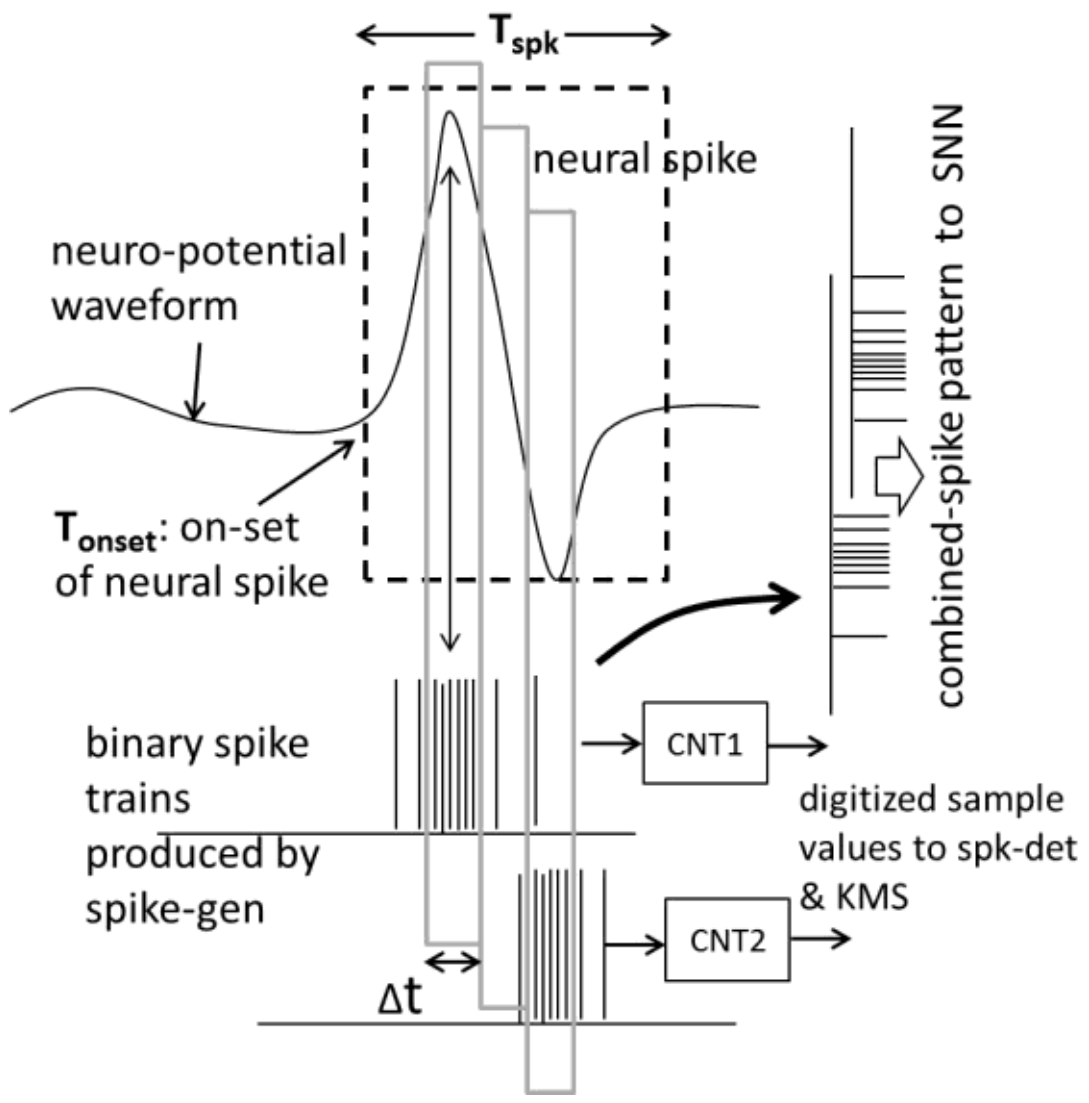

Fig. 2. N-spike encoding scheme for generating input feature vector to the SNN classifier.

## II. SYSTEM LEVEL DESCRIPTION OF NEURAL SPIKE SORTING MODULE

The system level diagram of the spike sorting module with a shared 2-step training module as proposed by Rakshit *et al.* is shown in Fig. 1 [19]. Each neuro-potential electrode is associated with a dedicated Neural-spike (N-spike) acquisition and sorting channel constituting of an Analog Front End (AFE) amplifier for pre-processing and filtering of the raw neural signal, a spike-generation block for encoding the neuro-potential waveform, an N-spike-detection block for capturing the onset of neural-spike and an SNN classifier.

By Neural-spike we imply the time-limited excursions in the recorded neuro-potential waveform, as shown in Fig. 2 (the box with dashed line). The occurrence of these N-spikes in the neuro-potential waveform needs to be detected and distinguished from spurious excursions due to noise [18] [27] [31]. These neural spikes are in general limited to few milliseconds in time and may constitute positive as well as negative excursions across the baseline [31]. In order to distinguish analog neural spike waveforms from binary spike produced by spike-gen block, the term N-spike is reserved for the prior.

The schemes for operating the spike-generation block in different modes for conversion of N-spike to its digitized form and binary spike trains and the N-spike-detection block for detecting the on-set of N-spike have been discussed earlier [19].

Each channel is selected in a sequential manner for the training of the binary spike trains. For a selected channel, firstly, the K-means block is used for labeling the N-spikes into different clusters. Secondly, the SNN trainer performs a supervised training on the labeled data and finally passes the computed trained weights through a de-multiplexer to the corresponding channel which is used by the SNN classifier to correctly predict the incoming binary spike trains. After the initial training performed by the K-means block, it is further triggered only when the distance between the initially trained means and the new N-spike entering the system keeps on exceeding a threshold value within a fixed span of time. Hence, learning occurs only when required which is very much necessary from the energy point of view as the K-means block would consume a significant portion of power during training.

## III. SHARED 2-STEP TRAINING MODULE

The shared 2-step training module consists of a K-means clustering block and SNN trainer. The training module is shared by the multiple channels functioning one at a time which is selected by a multiplexer. The input to the training module is the binary spike trains and the digitized N-spike and the output needed is the trained binary weight matrix formed using ‘1’s and ‘0’s which is fed back to the appropriate channel selected through a de-multiplexer. The function of the K-means block is to label the digitized N-spikes by categorizing them into different classes. These labeled N-spikes are then used by the SNN trainer block to perform a supervised training to generate the trained weight matrix. It is to be noted that re-training occurs only when the mismatch level of the new N-spikes with the trained means exceeds a threshold which is tracked by a conditional adaptation unit. In our case, we are classifying a dataset of a neural electrical signal having three classes which were taken from the database of Centre for Systems Neuroscience, University of Leicester [20][28].

### A. *Generation of the input features to the trainer*

The N-spike-detection module tracks the onset of N-spike based on the Non-linear Energy Operator (NEO) [19] [33]. Further, the spike generation module generates its downsampled digitized version when operated in mode 1 and the binary spike trains for mode 2. The time intervals having a higher magnitude of the amplitude of N-spike will generate a greater number of 1's as compared to the position where the N-spike amplitude is close to the zero level. The digitized N-spike of dimension $m \times n$ is used by the K-means trainer as input data. Here, m represents the downsampled features and n signifies the bit precision per sample. The binary spike trains consisting of 300 features, is used as input to the SNN trainer where each feature is either a `1' or `0' requiring a row in the crossbar network.

### B. *K-means trainer*

The K-means trainer, training the digitized N-spike collected in an SRAM memory through multiple channels is shown in Fig. 3. The major digital blocks in the data-path of the K-means trainer uses the stored data in the SRAM module. This is perhaps the most power consuming block in the system due to the involvement of memory access and training. The storage scheme of the digitized N-spikes through multiple electrode channels has been discussed earlier and the power consumption is in the range of ~13µW with an area ~0.33 mm$^2$ for a 16 channel system for a clock frequency of 8 KHz [32]. A point to be noted is that the dimension of m and n determines the limit of the number of N-spikes to be stored in the SPRAM module of dimension 8192×8. An optimum value of m=16 and n=5 has been considered which balances the trade-off between the accuracy in algorithmic level and power consumption in hardware implementation.

The K-means architecture is designed to have a predefined number of clusters (K=3) and a predefined number of iterations. In the K-means trainer architecture, the incoming digitized N-spike is stored in SPIKE DATA, the mean values are stored in MEAN1, MEAN2, MEAN3 and the total sum of data belonging to different classes for each iteration is stored in TOT1, TOT2, TOT3 which are 2-D registers. The dimension of the registers storing the digitized N-spike will be $m \times n$ whereas for the TOT registers $n$ will depend on the maximum number of data to be clustered.

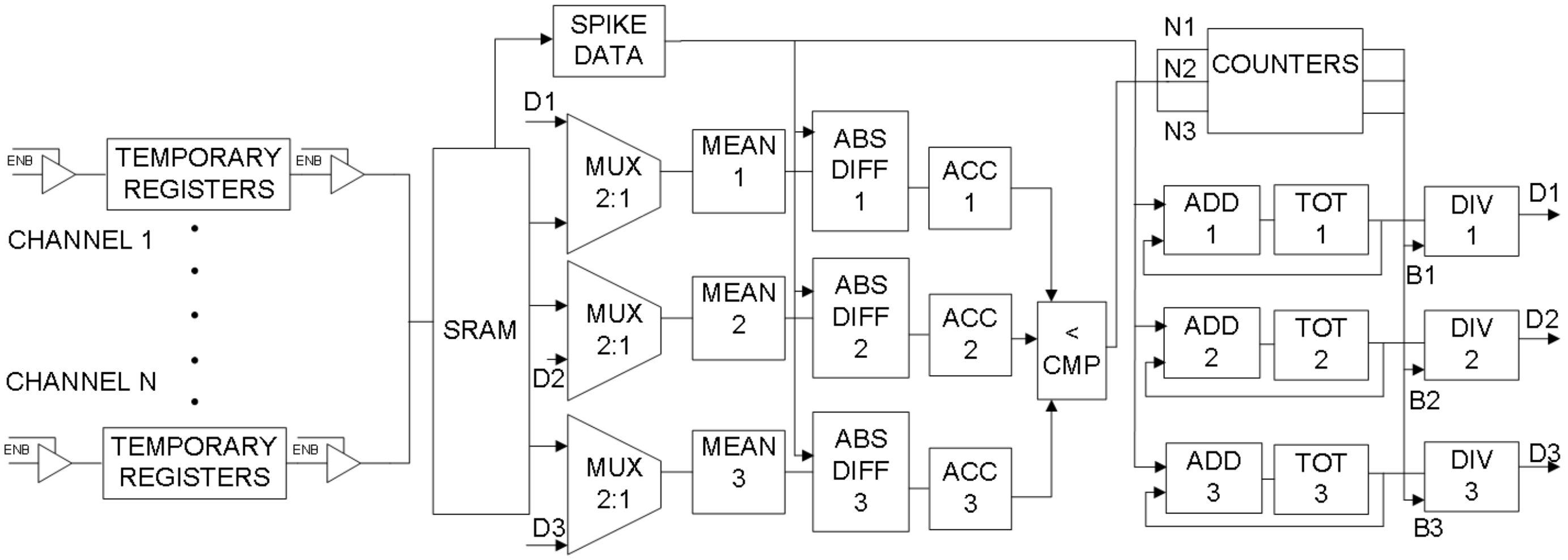

Fig. 3. Architecture of the K-means trainer depicting the major blocks involved in the data path for a multiple channel system.

TABLE I
SYNTHESIS REPORT OF K-MEANS RTL IN TERMS OF AREA AND POWER.

| Total area ($\mu m^2$) | Power consumption @ $f_{clk}$=500KHz ($\mu$W) | | | |
|---|---|---|---|---|
| | Internal | Switching | Leakage | Total |
| 177588 | 42.48 | 8.41 | 0.2353 | 51.12 |

Initially, three data are selected and initialized as the means in the respective blocks. For lower computational complexity, the hamming distance between an N-spike and the mean value is calculated instead of Euclidean distance metric using absolute difference (ABS DIFF) blocks for each sample which is then summed up by an accumulator (ACC) block. The minimum value at the output of the accumulators is checked through a less than comparator and the N-spike data is categorized. The count of the data (N1, N2, and N3) are stored in three different counters for every iteration. Meanwhile, three adders are used to add up the digitized spike data belonging to their respective class and store the sum of all the data into the corresponding TOT registers for every iteration. Finally, the new mean is calculated by dividing the total sum and count values (B1, B2, B3) of the respective classes through the divider (DIV) blocks. The previous value of mean contained in the MEAN blocks are then replaced by the new values of data (D1, D2, and D3) generated and then the process is repeated for the next iteration. The total power consumed by the K-means block when operated at $f_{clk}$=500 KHz is ~51μW in which the dominating component is the internal power consumption as mentioned in Table I.

### *C. SNN Trainer*

The input pattern to the SNN trainer is the binary spike trains representing the N-spike where each bit represents a neuron in the input layer and the required output generated is a binary weight matrix of dimension $N_I \times N_O$ (number of input neurons × number of classes).

The binary input data is trained using a modified version of Hebbian Spike Timing Dependent Plasticity (STDP) learning to obtain the trained binary weights [19] [21] [22]. The training process starts by feeding the input neurons with the $N_I$ (here, 300) bit spike pattern. The training algorithm of the supervised SNN is explained below and also depicted in Fig. 4.

1) Initially, the different inputs to the SNN which are in the form of binary spike trains (an equivalent form of an N-spike) are stored separately in their respective classes (in our case three classes) which had been determined by the K-means clustering block.

2) A fixed number (we have chosen five) of input is sent from each class in a sequential manner, to the SNN trainer, i.e. 5 inputs are sent from class 1, 2 and 3 respectively to the SNN and the process is repeated.

3) The weights are in the form of '1' or '0' format and the weight matrix of dimension ($N_I \times N_O$) is randomly initialized. Synaptic weight update may occur depending on the output neuron potential whenever a new input is received by the SNN network. Also, the number of data fed to the network is the number of iterations for weight updating.

4) The output neurons representing the number of classes are refreshed to its membrane resting potential after each weight update. Also, a leak constant is considered which reduces the output neurons potential in a uniform manner.

5) Whenever a new input is fed to the network, it affects the output neuron potential as given by equation (1).

TABLE II
SYNAPSE UPDATE PROBABILITTY FOR HEBBIAN/ ANTI-HEBBIAN STDP LEARNING.

| Presentation cycle | 1 | 2 | 3 | 4 | 5 |
|---|---|---|---|---|---|
| P_UP (0 to 1) | 8% | 6% | 4% | 2% | 0% |
| P_DN (1 to 0) | 0% | 0% | 0% | 0% | 8% |

$$V_j(t) = V_j(t-1) + \sum_{i=1}^{N} \left( W_{ij} \times A_i \right) - leak \qquad (1)$$

6) The weight update of the synapses connecting the winner output neuron with all the input neurons take place whenever the potential of the winner output neurons (OUT_MAX) exceeds the first threshold (thr1) value.

7) If OUT_MAX and the input data belong to the same class then the modified Hebbian STDP algorithm updates the weight matrix, otherwise, anti-Hebbian STDP algorithm is employed.

8) The second level of threshold (thr2) having a higher value checks whether the output neuron potential exceeds it. In that case, the update probabilities are reduced to prevent well-trained synapses from changing their states.

9) The probability of strengthening or weakening of a synapse depends on the presentation cycle of inputs from a particular class in accordance with the input firing pattern as depicted in Table II. The input neurons having a '1' value implies firing of neuron whereas the presentation cycle means the order number in which the input data from each class is provided to the SNN network. Earlier presentation cycle will have a higher chance of updating the synaptic weights and vice versa.

The SNN trainer would consume less power as compared to the K-means clustering block [19]. The power consumption of the SNN trainer as reported by Beinuo *et al.*is ~9.3μW/channel at 110 KHz clock frequency in 65 nm technology node [18]

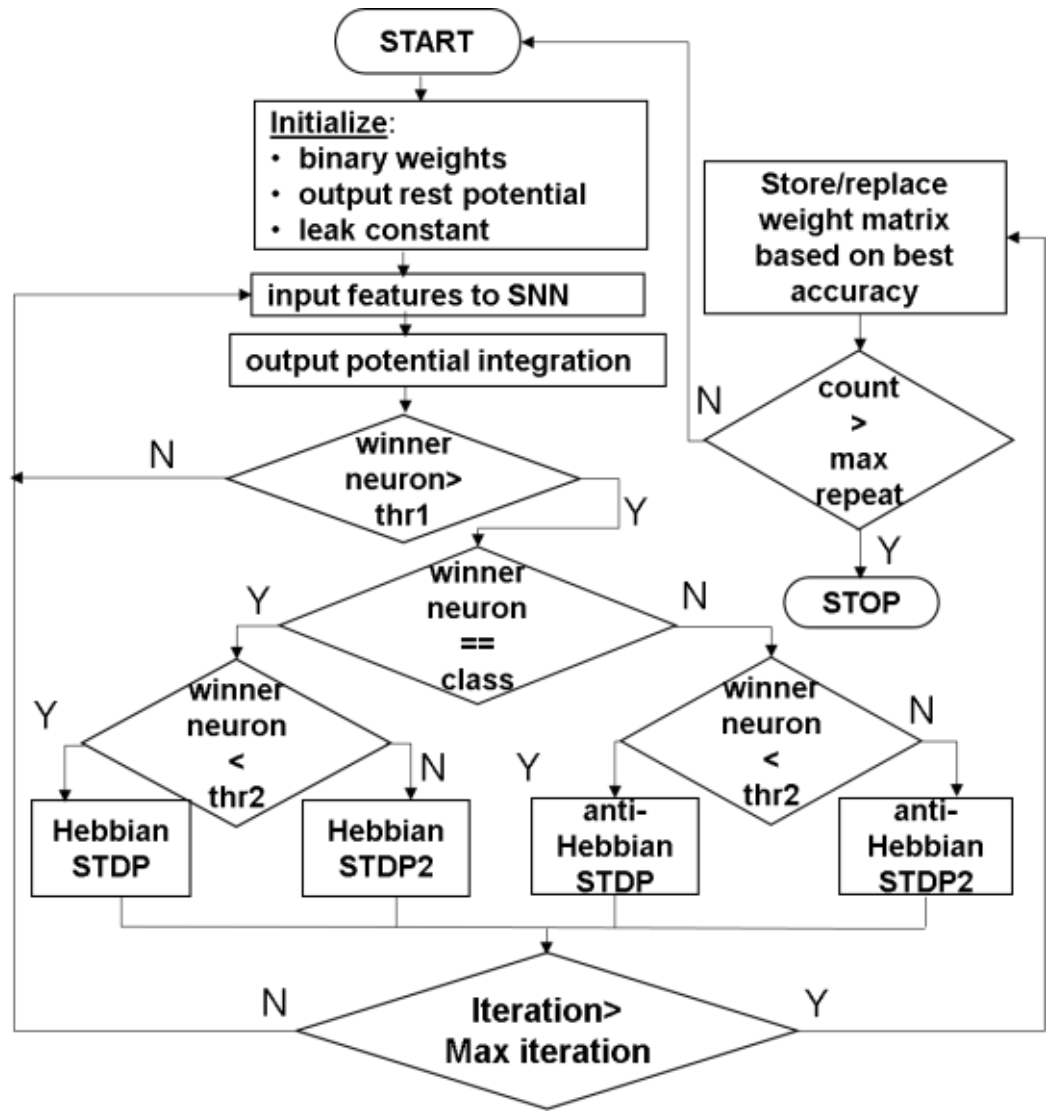

Fig. 4. Algorithm of SNN trainer with modified Hebbian STDP learning.

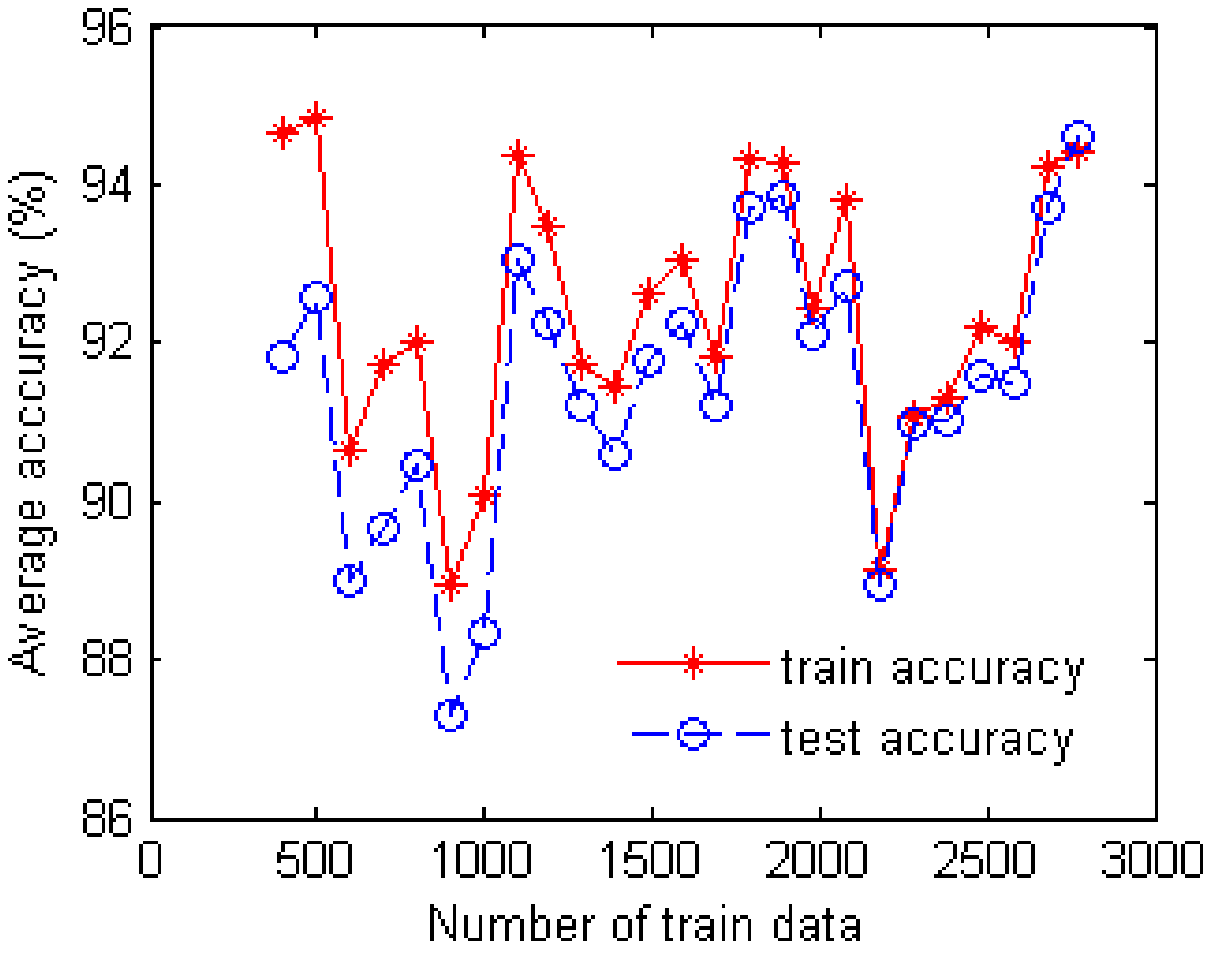

Fig. 5. Effect on average train and test accuracy with number of train data used for iterations.

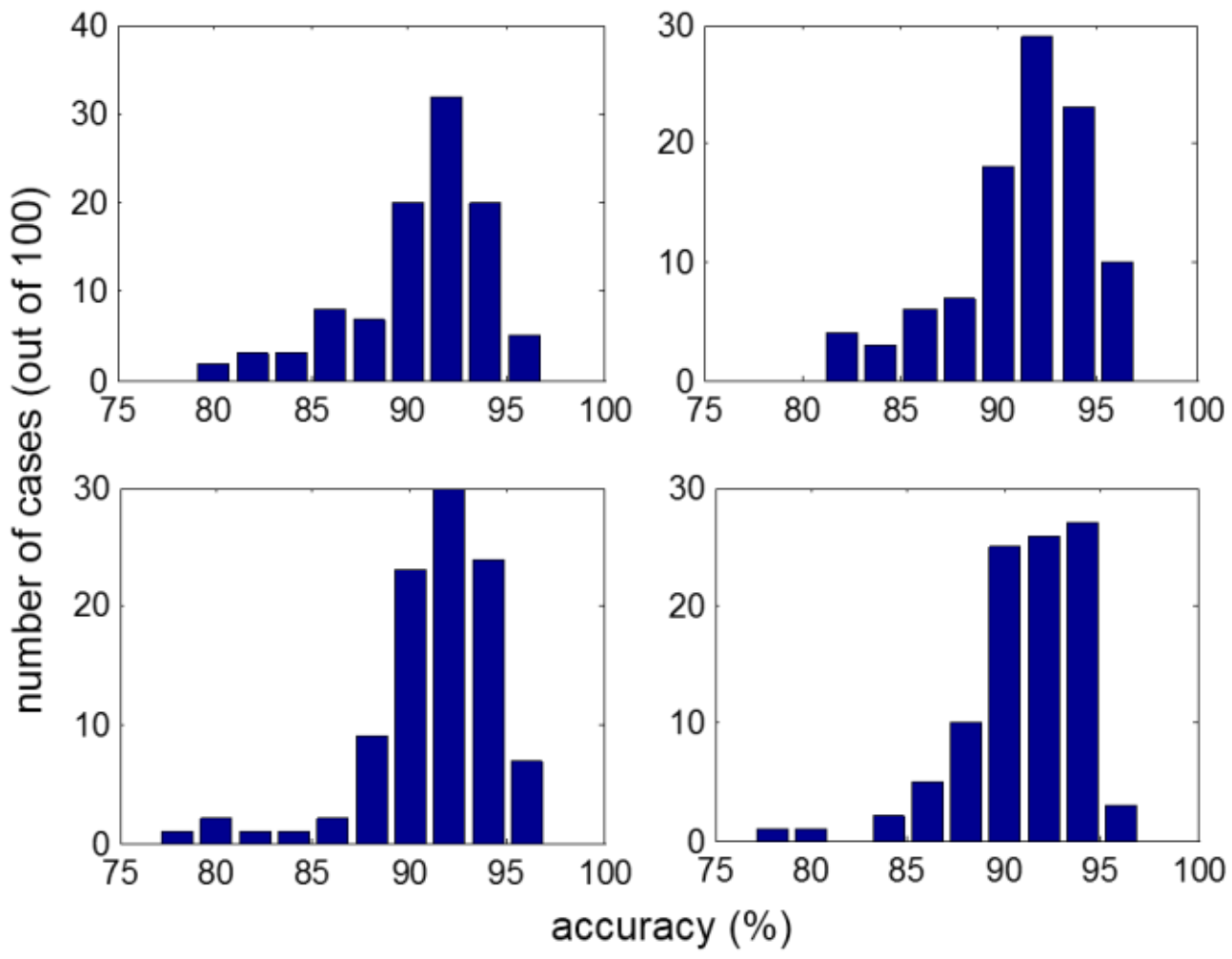

(a)

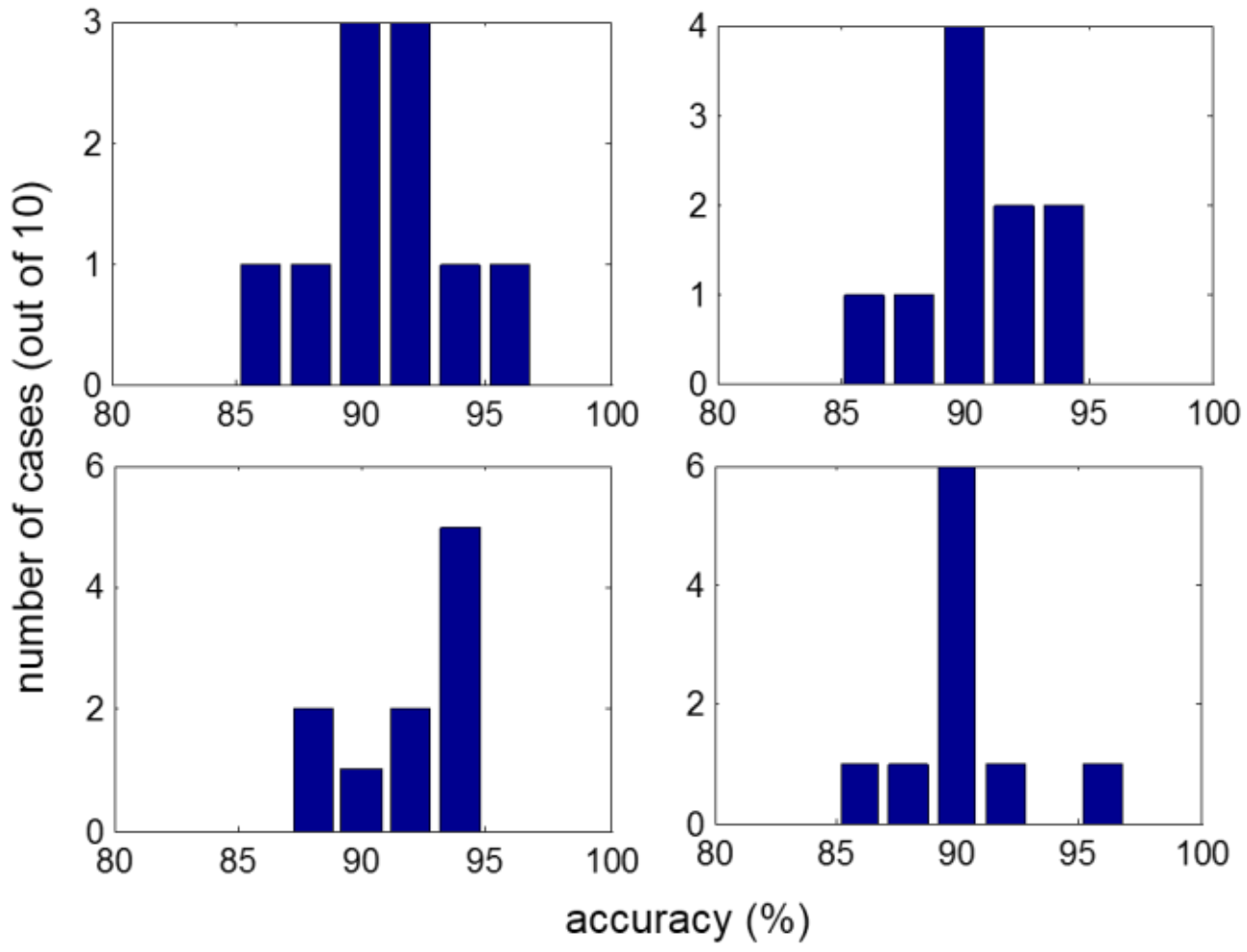

(b)

Fig. 6. Histogram for accuracy levels for (a) 100 runs (b) 10 runs for 4 different simulation runs.

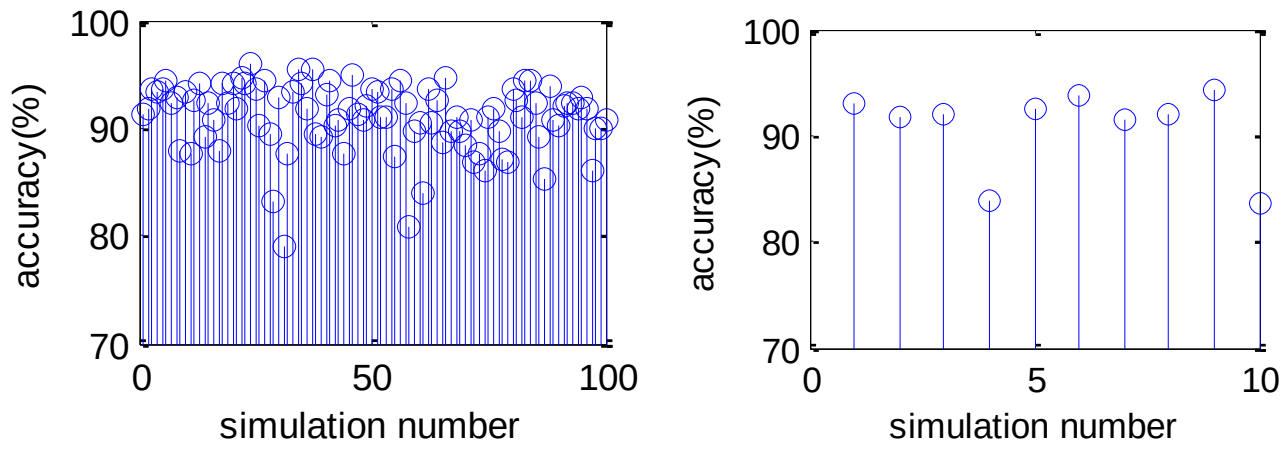

(a) (b)

Fig. 7. Variation in train accuracy for (a) 100 (b) 10 different consecutive runs.

The working of the SNN network was analyzed using 3410 N-spike data having three different classes, each represented by $N_I$ (300) binary bits. The total dataset was divided into the train and test data and the average train/test accuracy of the network versus the number of trained data used for training the network is shown in Fig. 5. The rest of the N-spike were used as test data to measure the accuracy. It can be clearly understood that unlike other neural networks in which the training accuracy improves with more number of train data, there lies a fluctuation in the accuracy due to randomness involved in the training process in the form of probability of updating synapses. Also, the difference in train and test accuracy is almost negligible after a certain number of train data (>1500 in this case). Hence, the number of train data (iterations) can be fixed as a constant value when thinking from the hardware perspective. But due to the fluctuation of the accuracy level within a certain acceptable range, the SNN network is required to be trained repeatedly for a fixed number of times and trained weights should be considered for the best case.

In order to determine the number of times the SNN trainer should be operated (max repeat), it is first simulated 100 times for a particular case (with 1800 train data) and the histogram for the accuracy level is plotted for 4 different times as shown in Fig. 6 (a). Next, the number of runtimes is reduced to 10 and the same is repeated as shown in Fig. 6 (b). By comparing the two cases, it is quite clear that the probability of getting a higher accuracy is similar even when the number of runtimes is reduced to 10 which would be very much appreciable from the hardware perspective in terms of computation complexity. Also, in Fig. 7 train accuracy for 100 and 10 different runs are plotted and the case with the best accuracy is considered.

## IV. Mathematical Model of a Memristive Crossbar Network

### A. *Architecture of a memristive crossbar network*

The crossbar network constitutes of memristors (e.g. Ag-Si) with conductance $G_{ij}$ (resistance $R_{ij}$) interconnecting two sets of metal bars ($i_{th}$ horizontal bar and $j_{th}$ in-plane bar) with parasitic resistance $R_p$ (conductance $G_p$) between two nodes due to metal wires. The terminal ends of the in-plane bars are connected to an analog device having an input resistance of $R_t$ (conductance $G_t$) as shown in Fig. 8.

In this architecture, the input features are converted to its current equivalent and fed at the input of the horizontal bars. The memristor represents the trained weights. The number of vertical columns is taken to be one more than the number of categories the entire dataset is to be classified. The extra column acts as a dummy column which is required to equalize the total conductance in each row of the network to maintain uniformity and proper operation of the network. Hence, area density occupied by the memristive crossbar depends on the number of bits needed to form the weight matrix plus the number of bits needed to allocate the additional dummy column.

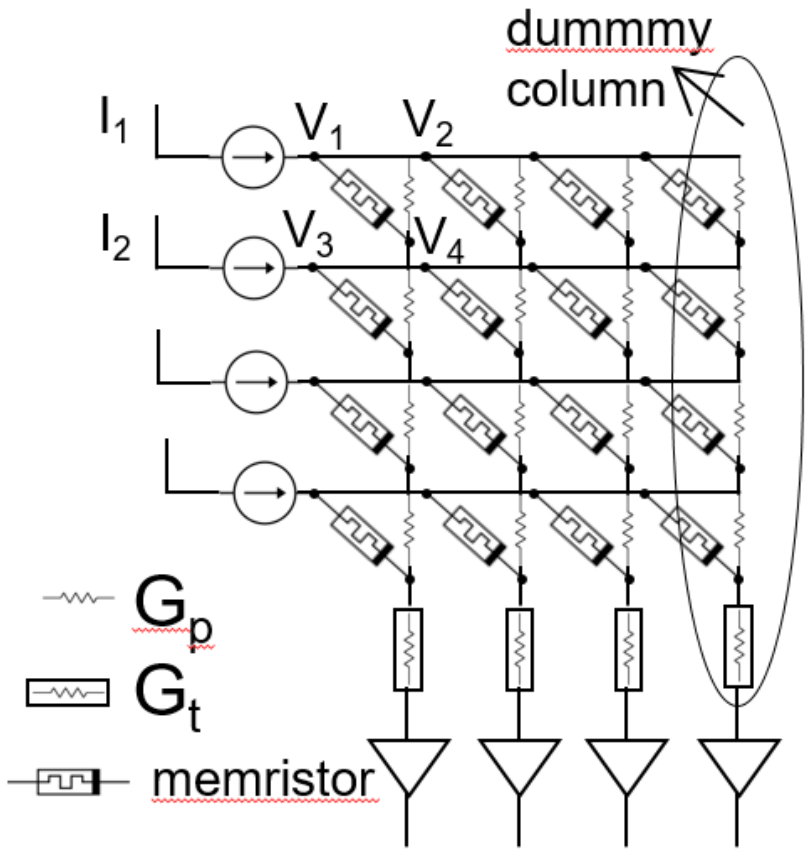

Fig. 8. A schematic of a 4 x 4 memristive crossbar network for three classes including a dummy column.

### B. *Computing with the memristive crossbar network*

The resistive crossbar network is used to classify data from a dataset when the feature size of individual data, trained weights and the number of classes to which the data can belong is known. It can classify data efficiently which is very much comparable to the accuracy of classification achieved at the algorithmic level, provided the circuit parameters are chosen appropriately. In general, this crossbar architecture can be used to classify any kind of datasets if the features extracted are in the form of numbers. In our case, we have considered the data set to be the binary spike trains. The node voltages and the corresponding current flowing through it can be calculated mathematically by solving KCL and could be represented as given in equation (2)

$$[G]_{N\times N}\, N[V]_{N\times 1} = [I]_{N\times 1} \qquad (2)$$

Here, G, V and I represent the conductance, voltage and current matrix formed by solving KCL at all the nodes present in the network while taking the effect of parasitic conductance and termination conductance into account. Here, *N=2×row×column* determines the size of the conductance matrix. The [G], [V] and [I] matrices in expanded form for a 2×2 crossbar can be represented as follows:

$$[G]=\begin{bmatrix} A & -G_{11'} & 0 & 0 & -G_p & 0 & 0 & 0 \\ -G_{11^1} & A & 0 & -G_p & 0 & 0 & 0 & 0 \\ 0 & 0 & B & -G_{22'} & 0 & 0 & -G_p & 0 \\ 0 & -G_p & -G_{22'} & C & 0 & 0 & 0 & 0 \\ -G_p & 0 & 0 & 0 & D & -G_{33'} & 0 & 0 \\ 0 & 0 & 0 & 0 & -G_{33'} & D & 0 & -G_p \\ 0 & 0 & -G_p & 0 & 0 & 0 & E & -G_{44'} \\ 0 & 0 & 0 & 0 & 0 & -G_p & -G_{44'} & F \end{bmatrix}$$

$$[V]=\begin{bmatrix} V_1 & V_1' & V_2 & V_2' & V_3 & V_3' & V_4 & V_4' \end{bmatrix}^T$$

$$[I]=\begin{bmatrix} I_1 & 0 & I_2 & 0 & 0 & 0 & 0 & 0 \end{bmatrix}^T$$

In the above expression, A= ($G_{11'}$+$G_p$), B= ($G_{22'}$+$G_p$), C= (B+$G_t$), D= ($G_{33'}$+$G_p$), E= ($G_{44'}$+$G_p$), F= (E+$G_t$), in which $G_p$ and $G_t$ are the parasitic conductance at the junction node and the termination conductance at the output node of the crossbar network. Also, $G_{11'}$, $G_{22'}$, $G_{33'}$ and $G_{44'}$ represent the conductance of the memristor used to connect the horizontal and vertical lines of the crossbar network. $V_1$, $V'_1$, $V_2$, $V'_2$, $V_3$, $V'_3$, $V_4$, $V'_4$ and $I_1$, $I_2$ represent all the node voltages and the input current provided to the crossbar network.

## V. SPIKING NEURAL NETWORK (SNN) CLASSIFIER ARCHITECTURE

Architecture for the SNN classifier can be implemented either in a fully digital fashion using digital blocks or at the transistor level using simple analog circuitry. A comparison between the two schemes has been presented in this paper.

### A. *Digital implementation of SNN Classifier*

The SNN architecture implemented digitally comprises of digital blocks like basic gates, counters, and comparator for determining the output class of the examined N-spike at the input as shown in Fig. 9. The number of counters is determined by the number of classes considered and the number of bits of each counter depends on the number of input neurons considered during training. The trained network considered is of dimension $N_I$×$N_O$, wherein $N_I$ and $N_O$ are the number of input and output neurons. The use of a shift register consisting of $N_I$ number of flip flops is to select each indices of the weight matrix one by one and multiply it with the incoming feature of each N-spike. Meanwhile, $N_O$ number of counters representing the output neurons keep on incrementing according to the input spike bit and trained weight values, which is further fed to a greater than comparator to determine the winner neuron at the time when all the $N_I$ features of a particular N-spike is fed into the network. Though the digital implementation of the SNN classifier is simpler and robust it would be consuming a significant amount of power.

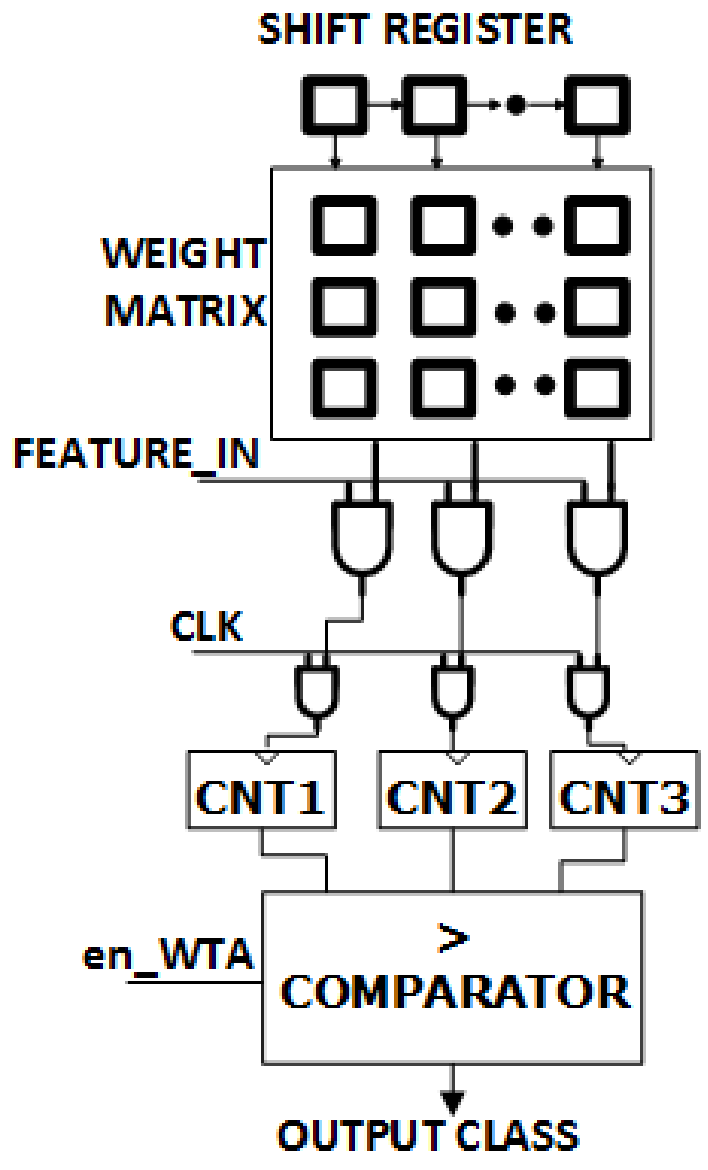

Fig. 9. Schematic of SNN Classifier using digital gates [19].

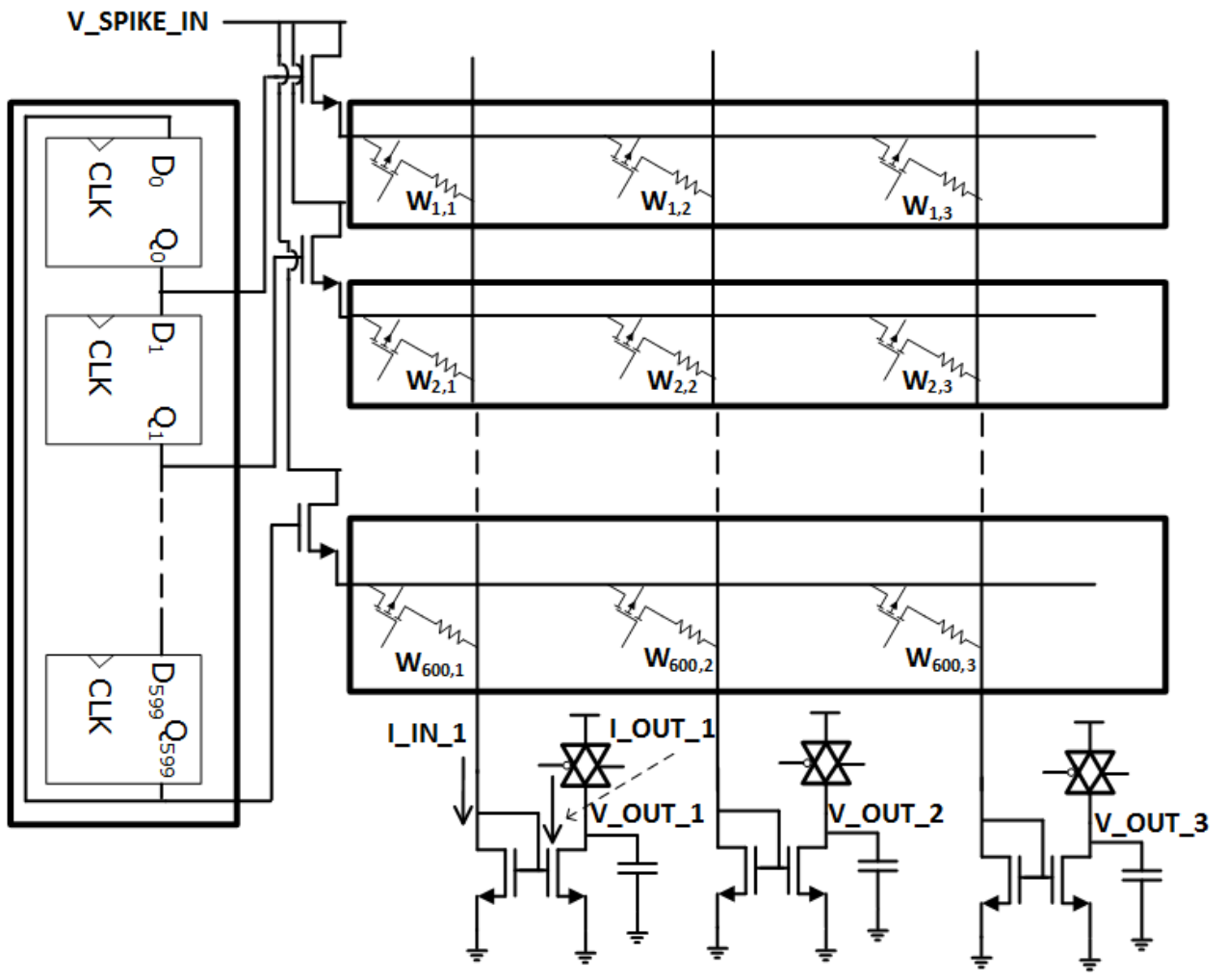

Fig. 10. Schematic of the SNN Classifier in a memristive crossbar architecture fashion.

## *B. Low power analog implementation*

The SNN classifier can also be implemented in a power and area efficient way using simple analog circuits in a crossbar type architecture as shown in Fig. 10. The incoming features of the N-spikes are denoted by V_SPIKE_IN. The number of rows and columns in the crossbar network is equal to the number of features present in the N-spike and the number of classes available. A SISO shift register is used for controlling the selection of a particular row at each clock cycle. Hence, only after $N_I$ clock cycles, an entire N-spike data is completely fed into the system. One important point to be considered which is not shown in the figure is the inclusion of dummy columns whose role is for equalizing the total resistance at each row. Also, each memristor shown in the crossbar network is emulated using a PMOS transistor and a resistor. Whenever the PMOS switch is in OFF state the memristance is high representing a low value of the weight. The equivalent value of the trained weights is applied at the gate terminal of the PMOS transistors used which modulates the resistance of the synaptic junction to a high and low value corresponding to the '0' and '1' state. Initially just before the entry of an N-spike data, the capacitor at the output node of the current mirror shown in Fig. 10 is charged to $V_{DD}$ through a transmission gate switch. Whenever a row is activated, the corresponding feature is transmitted through the synaptic weights in the form of an equivalent current which is further copied by the current mirror circuitry thereby causing discharge of the node capacitance by a small amount. When all the features of an N-spike are computed the final value of the node voltages obtained are compared to determine the class to which the N-spike belong.

The capacitance value to be used at the output node of each column can be determined from equation (3) in which $C_L$, $\Delta V$, N, $I_{synapse}$, $T_{on}$ are the capacitance, maximum voltage drop across the capacitance, number of features representing '1' in the input N-spike, the synaptic current mirrored to the current mirror and time duration for activating the rows for passing the input features respectively.

$$C_L \Delta V = N \times I_{synapse} \times T_{on} \tag{3}$$

One important point to note regarding $I_{synapse}$ is that if it is too low (<100nA) then a little variation in the gate voltage at the PMOS synapses will cause a larger fluctuation in its resistance value as the MOS device has an exponential dependency on the drain to source voltage in the subthreshold regime of operation. Hence the synaptic PMOS should not be operated in the deep subthreshold region of operation. On the other hand, if the higher value (>1μA) of synaptic current is allowed then the power dissipation of the overall circuitry will increase. Hence the input bit patterns entering the system are allowed for a short period of time ($T_{on}$) while allowing an optimum value of synaptic current through it as shown in Fig. 13(b). If $T_{on}$ is chosen to be 1% of the data bit period ($T_{data}=1/f_{data}$) then it can be expressed as $T_{on}=0.01/f_{data}$.

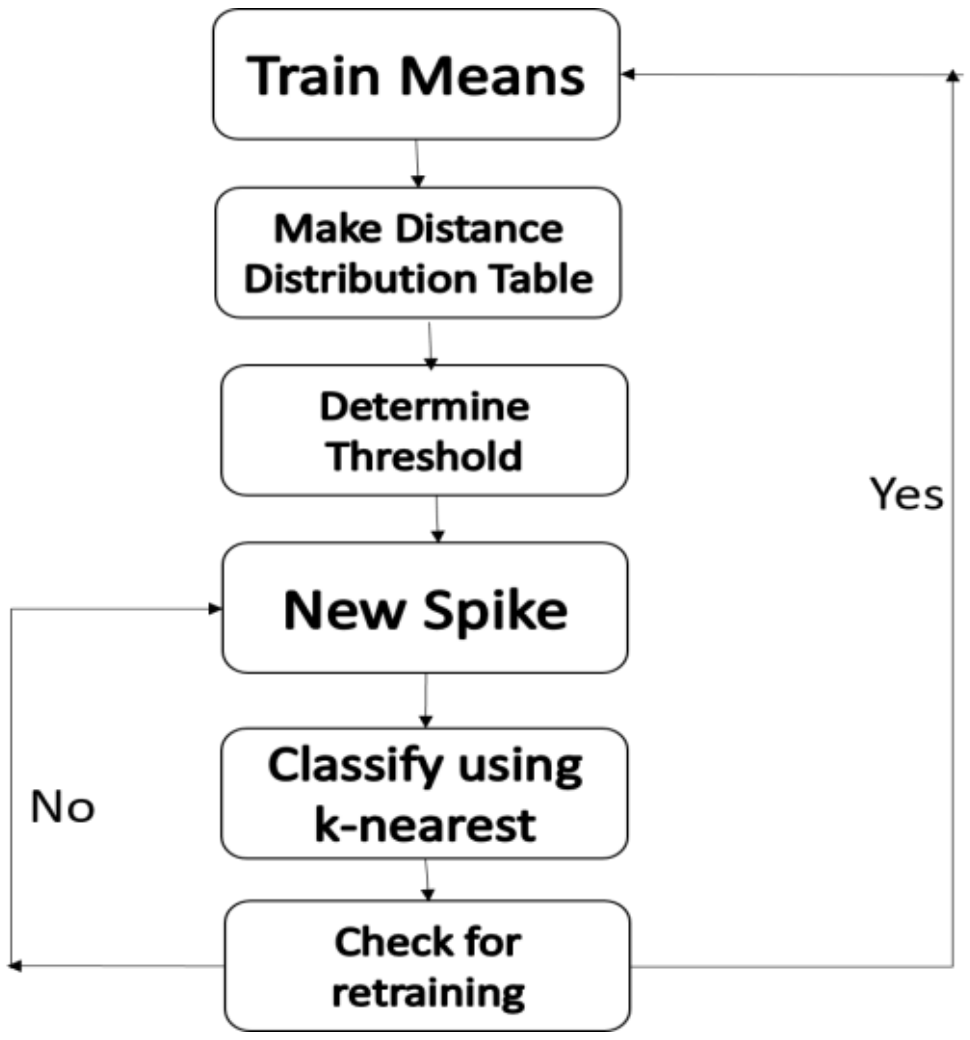

Fig. 11. Flowchart for retraining of the k-means classifier based on conditional adaptation.

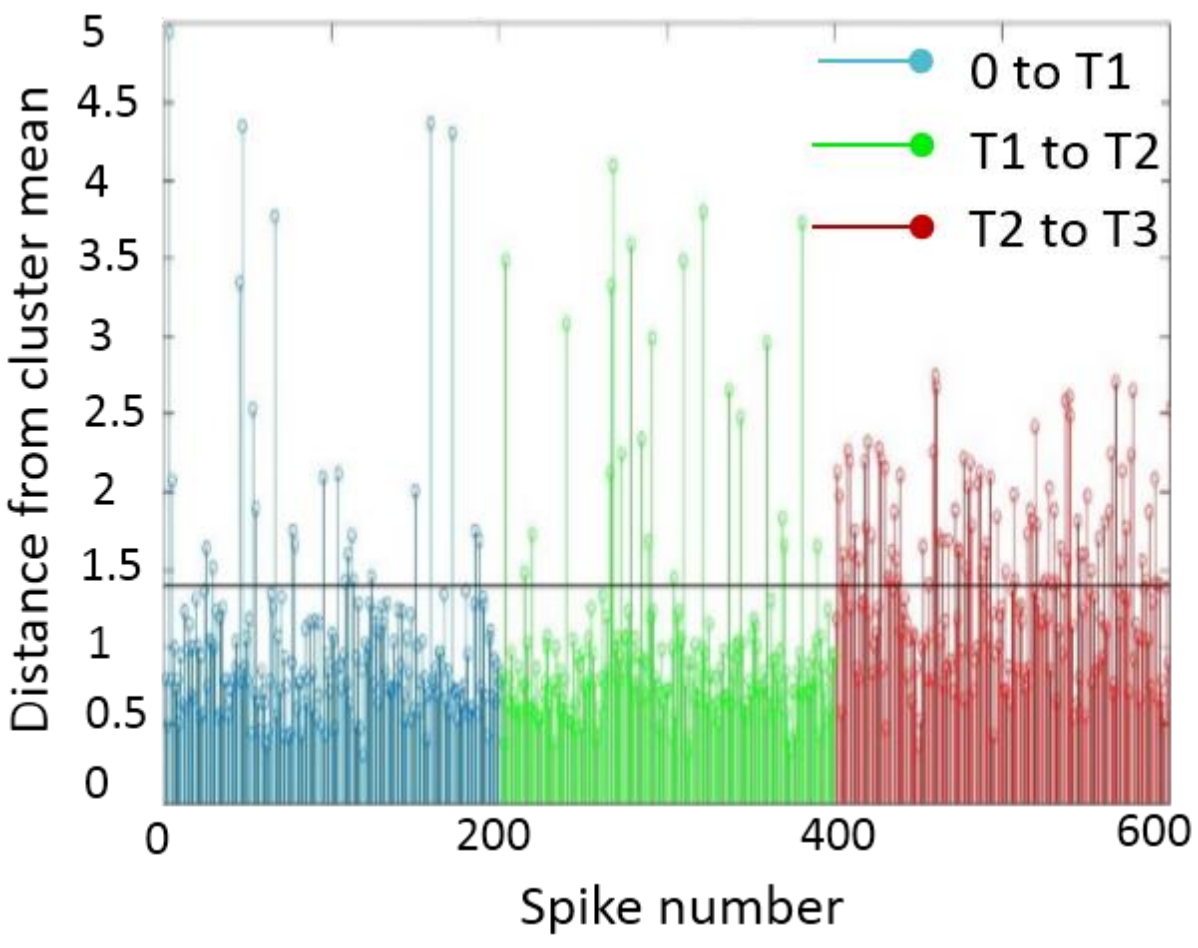

Fig. 12. Temporal Distribution of Spike Distance from Cluster Mean

## VI. ADAPTATION

### A. *Detection of mismatch*

The output node voltage of the crossbar circuitry is compared with a reference threshold voltage of a fixed number of N-spike samples. If the output node voltage crosses the reference threshold voltage on a frequent basis, then a mismatch is detected and the incoming N-spike does not belong to any of the predefined classes. The situation is tackled by retraining of the network with the inclusion of the newer N-spikes for generating a new set of trained weights.

### B. *Retraining*

The flowchart of the algorithm of retraining the K-means classifier is shown in Fig. 11. Initially, a set of data is used for calculating the means of the different clusters. A distance distribution table is made which plots the distance of each trained data with the mean value of the cluster to which it belongs. Then the threshold value of the distance is estimated based on some criteria from the distance distribution table. Whenever a new spike data enters the system, it is classified using K-nearest algorithm to its respective cluster. If the distance of the new spike data from the cluster mean exceeds the determined threshold value, then the k-means classifier is required to be retrained.

### C. *Need for conditional adaptation*

In Fig. 12 the temporal distribution of distance of spike from cluster mean is shown. It is divided into three temporal sections: 0 to T1, T1 to T2 and T2 to T3. The section 0 to T1 consists of spikes used for training of the means, T1 to T2 consists of spikes similar to original spike waveform and T2 to T3 consists of spikes different from original waveform. It could be visualized from section T2 to T3 that the spikes have a distance to the cluster mean greater than the threshold on a more frequent basis and this will trigger the retraining.

## VII. RESULTS AND DISCUSSION

### A. *Comparison between digital and analog implementation of Spiking Neural Network Classifier*

The implementation of the SNN classifier was done at a clock frequency of 200 KHz which depends on the bit duration (5µs) of the features of the incoming N-spikes. From Fig. 9 it is seen that a memory matrix is required to store the trained weights which are accessed at every clock cycle thereby contributing to the majority of power consumption. Apart from the memory module other sequential and combinational circuit elements also dissipate a significant amount of energy. The SNN classifier excluding the memory power consumes an average power of ~300nW at 200 KHz clock frequency as estimated in 180nm CMOS process technology using Synopsys Design Compiler tool.

On the other hand, in the crossbar type implementation, accessing the trained weights stored in a memory takes place only when there is an updating in the weights, which are then applied at the gate terminal of the PMOS synaptic switches. The majority of power consumed in this type of implementation is due to the SISO shift register which transmits a ‘1’ at every clock for selecting the rows in a sequential manner. The average power consumption per cycle for the crossbar architecture can be estimated using the following equation (4),

$$P_{avg} = I_{synapse} \times n_{column} \times V_{DD} \times \beta \tag{4}$$

Here, $I_{synapse}$, $n_{column}$ are the synaptic current (~0.3µA) flowing through the synaptic junction and the average number of columns used in the crossbar network. $V_{DD}=1.2V$ and $\beta = \frac{\Delta t}{f_{data}} = \frac{50e-9}{5e-6} = 0.01$ are the supply voltage and the fraction of on duration time of the incoming N-spike bit respectively. Hence, the average power consumed is ~10nW. The average power consumed by the SISO shift register transmitting a ‘1’ at every clock cycle is ~5.2µW. The accuracy of the crossbar architecture was found to be ~96% when compared to the algorithmic level for 1200 test data.

### B. *Effect of variation in memristor in SNN classifier*

The effect on train/test accuracy with variation in memristance value is analyzed. Random variation of up to 200% in the binary

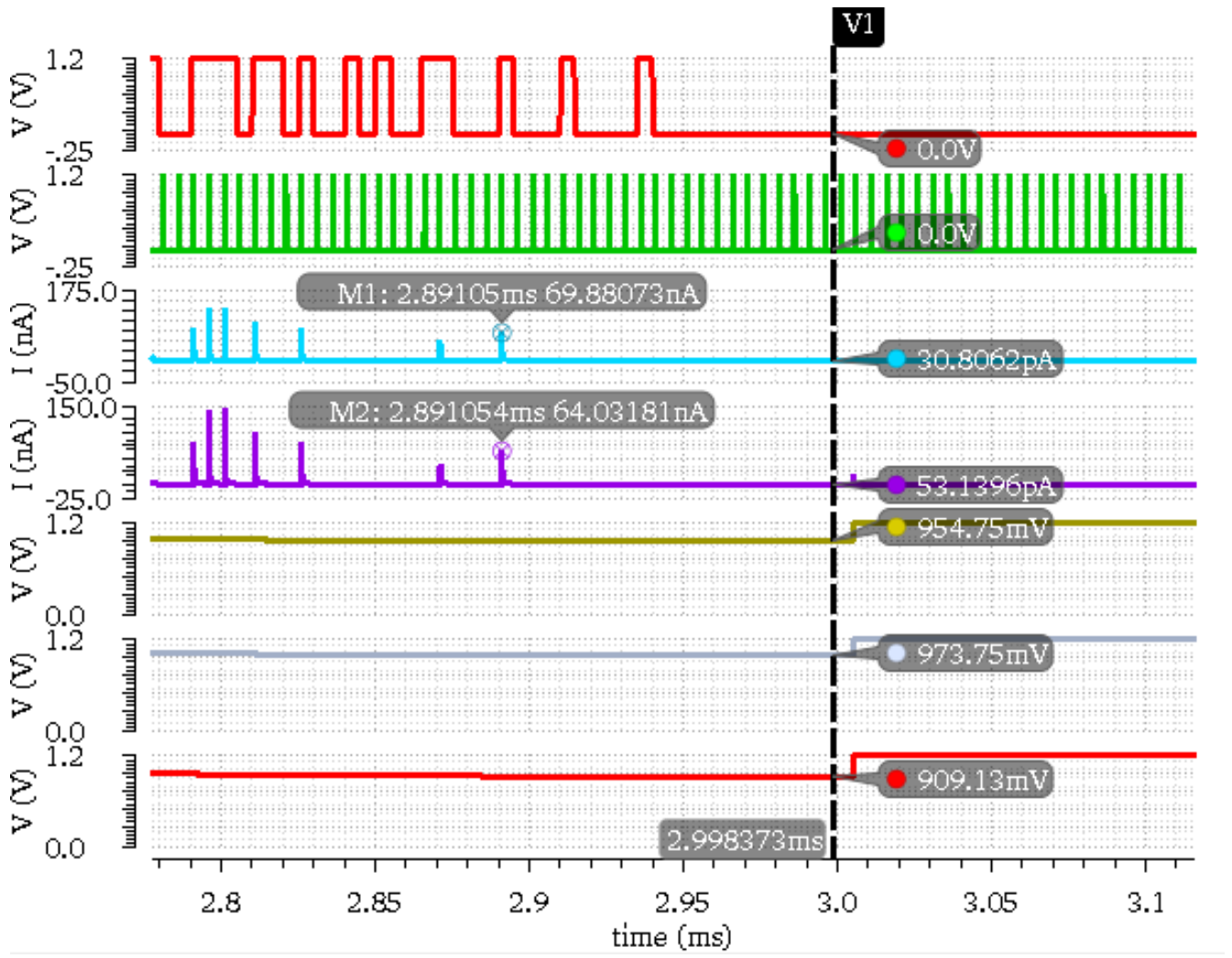

(a)

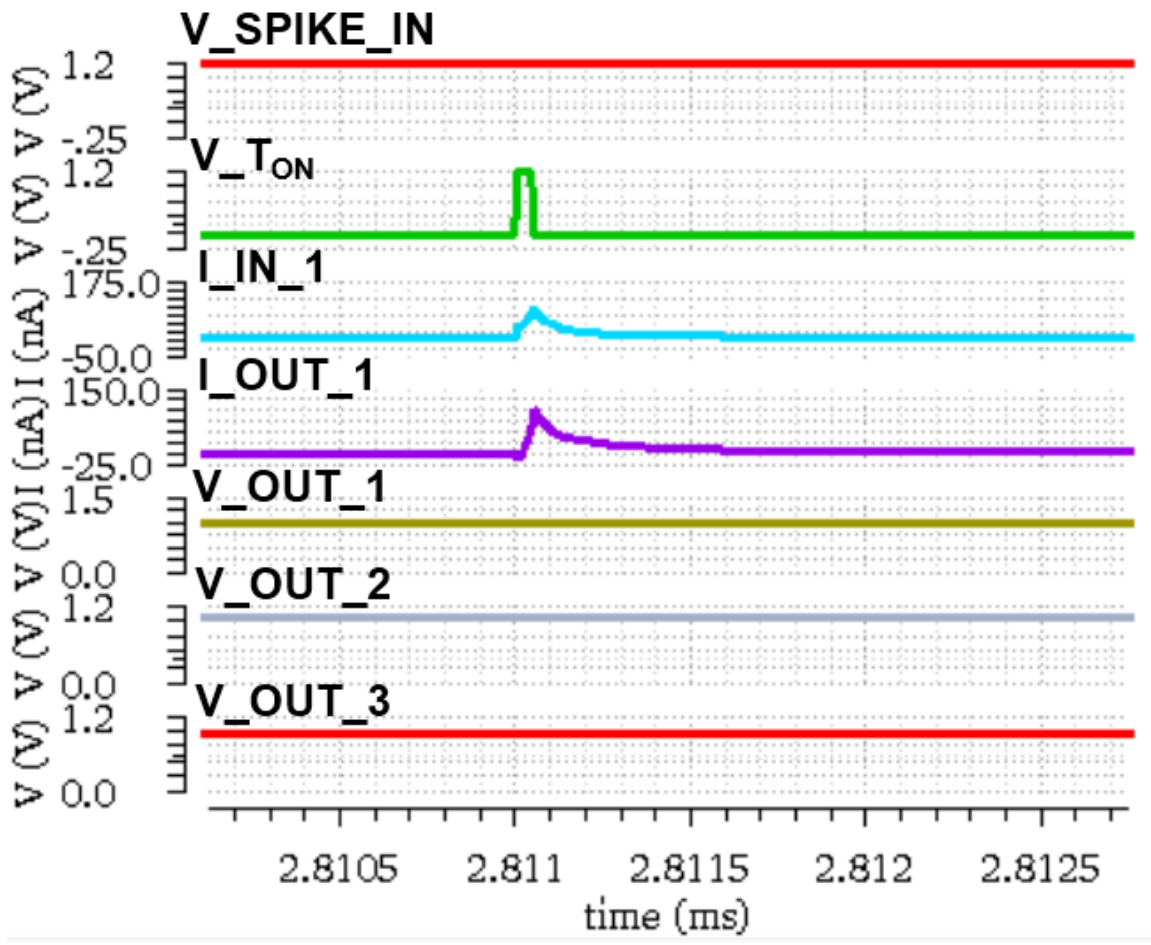

(b)

Fig. 13. Timing diagram of the crossbar architecture depicting: (a) the final output node voltage of the columns in the crossbar network for a N-spike input, (b) zoomed portion showing the current copy mechanism.

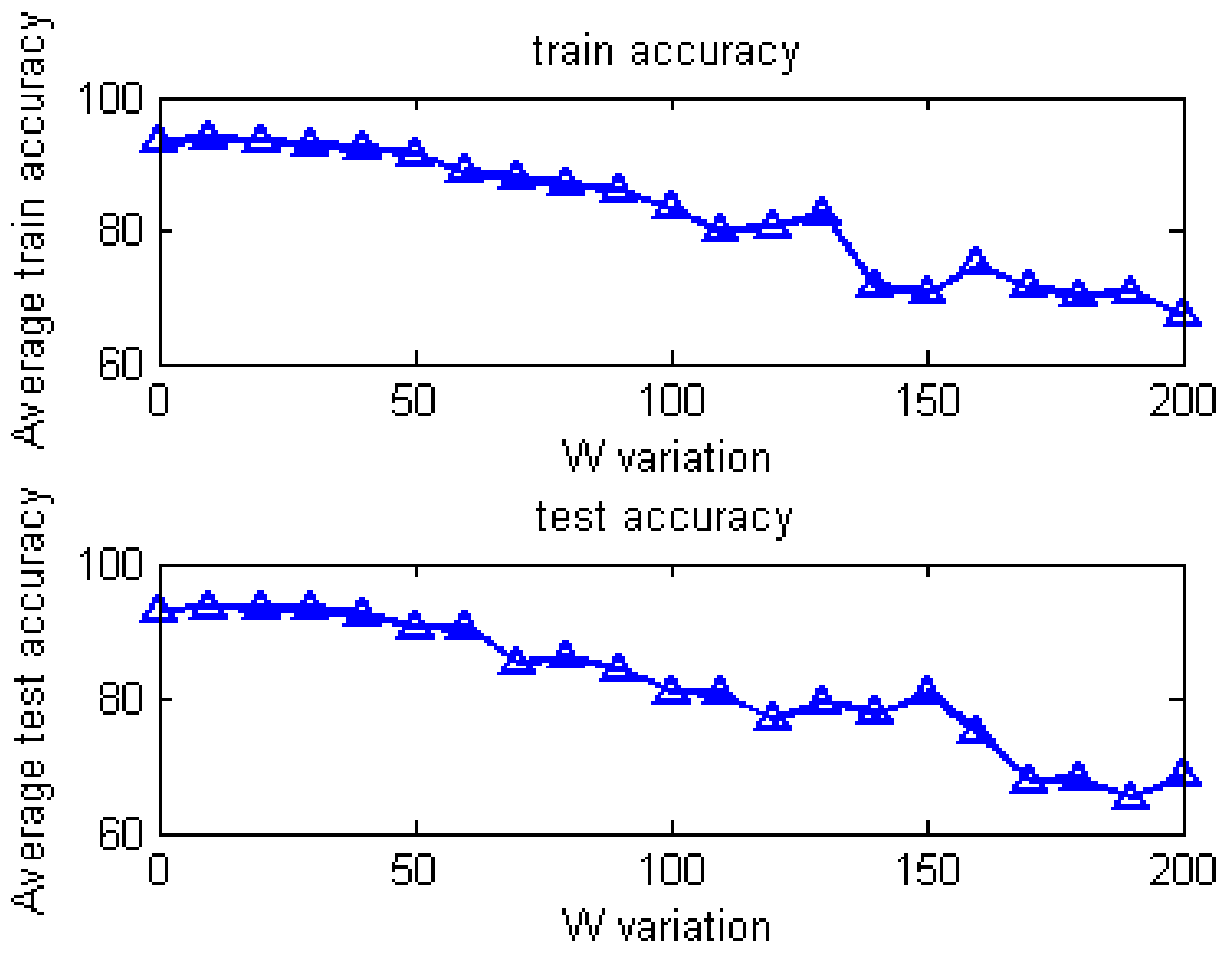

(a)

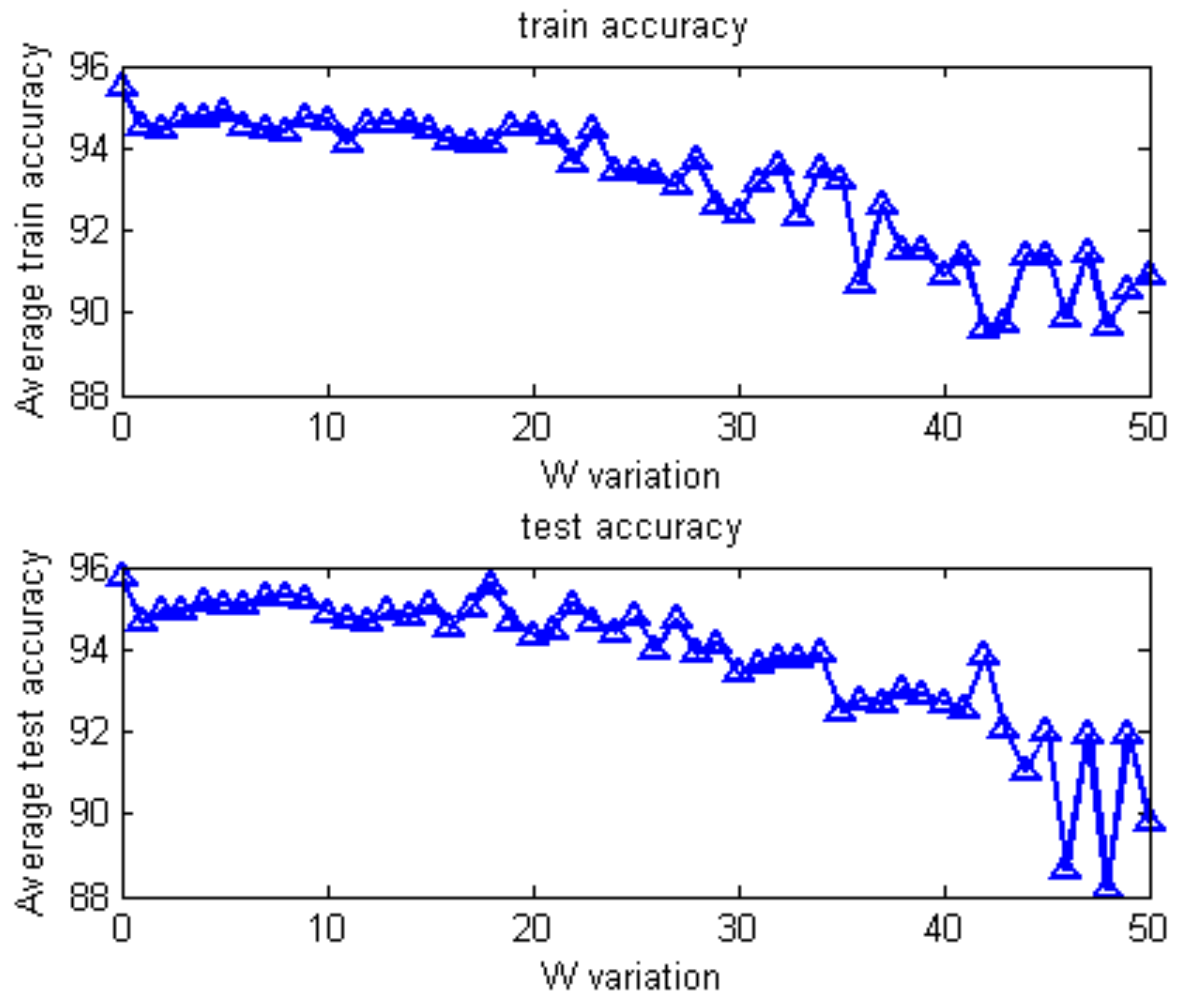

(b)

Fig. 14. Effect on average train/test accuracy with variation in memristance values (trained weights) for:

(a) Maximum random variation of 200%.
(b) Maximum random variation of 50%.

trained weights is incorporated and the drop in average train/test classification accuracy is plotted as shown in Fig. 14 (a). Also, to get a better understanding of the effect of variation, more number of points were considered for a maximum random variation of 50% as shown in Fig. 14(b). It is understood that the accuracy level falls gradually with increase in variation of memristance value and starts degrading at a steeper rate only when the variation is greater than 30%. Hence, the accuracy will not degrade significantly for small amount of PVT (process, temperature and voltage) variations of the memristor when implemented at the circuit level.

## VIII. Conclusion

The power consumption of the major sub-blocks used in the 2-step shared training module of the proposed spike sorting system are discussed. It is understood that major power consumption is due to memory access and K-means trainer. Also, a retraining technique using K-means based conditional adaptation is discussed for re-training of the network whenever necessary.

A comparison has been done on the implementation of a Spiking Neural Network (SNN) classifier in hardware between fully digital implementation and memristor-based crossbar network. The power consumption of the later is significantly lower compared to the former case due to flexibility in choosing design parameters at the transistor level. It is also understood that the for variations less than 30% in memristance due to PVT fluctuations would not affect the accuracy by a large amount indicating the robustness of the crossbar SNN classifier architecture scheme.

## References

[1] Mead, Carver. "Neuromorphic electronic systems." Proceedings of the IEEE, vol. 78, no. 10, pp. 1629-1636, 1990.
[2] Indiveri, Giacomo, and Timothy K. Horiuchi. "Frontiers in neuromorphic engineering." Frontiers in neuroscience vol. 5, 2011.
[3] Ebong, Idongesit E., and Pinaki Mazumder. "CMOS and memristor-based neural network design for position detection." Proceedings of the IEEE, vol. 100, no. 6, pp. 2050-2060, 2012.
[4] Chua, Leon. "Memristor-the missing circuit element." IEEE Transactions on circuit theory 18.5 (1971): 507-519.
[5] Jo, Sung Hyun, et al. "Nanoscale memristor device as synapse in neuromorphic systems." Nano letter, vol.10, no.4, pp. 1297-1301, 2010.
[6] Sheri, Ahmad Muqeem, et al. "Neuromorphic character recognition system with two PCMO memristors as a synapse." IEEE Transactions on Industrial Electronics 61.6 (2014): 2933-2941.
[7] Park, Sangsu, et al. "Electronic system with memristive synapses for pattern recognition." Scientific reports 5 (2015): 10123.
[8] Chu, Myonglae, et al. "Neuromorphic hardware system for visual pattern recognition with memristor array and CMOS neuron." IEEE Transactions on Industrial Electronics 62.4 (2015): 2410-2419.
[9] Starzyk, Janusz A. "Memristor crossbar architecture for synchronous neural networks." IEEE Transactions on Circuits and Systems I: Regular Papers 61.8 (2014): 2390-2401.
[10] Kim, Hyongsuk, et al. "Memristor bridge synapses." Proceedings of the IEEE 100.6 (2012): 2061-2070.
[11] Jo, Sung Hyun, Kuk-Hwan Kim, and Wei Lu. "High-density crossbar arrays based on a Si memristive system." Nano letters 9.2 (2009): 870-874.
[12] Maass, Wolfgang. "Networks of spiking neurons: the third generation of neural network models." Neural networks 10.9 (1997): 1659-1671.
[13] Vreeken, Jilles. "Spiking neural networks, an introduction." (2003).
[14] Paugam-Moisy, Hélene, and Sander Bohte. "Computing with spiking neuron networks." Handbook of natural computing. Springer Berlin Heidelberg, 2012. 335-376.
[15] Andrew, Alex M. "Spiking neuron models: Single neurons, populations, plasticity." Kybernetes 32.7/8 (2003).
[16] Izhikevich, Eugene M. "Simple model of spiking neurons." IEEE Transactions on neural networks 14.6 (2003): 1569-1572.

[17] Abbott, Larry F. "Lapicque's introduction of the integrate-and-fire model neuron (1907)." Brain research bulletin 50.5 (1999): 303-304.
[18] Zhang, Beinuo, et al. "A neuromorphic neural spike clustering processor for deep-brain sensing and stimulation systems." Low Power Electronics and Design (ISLPED), 2015 IEEE/ACM International Symposium on. IEEE, 2015.
[19] Pathak, Rakshit, et al. "Low Power Implantable Spike Sorting Scheme Based on Neuromorphic Classifier with Supervised Training Engine." VLSI (ISVLSI), 2017 IEEE Computer Society Annual Symposium on. IEEE, 2017.
[20] http://www2.le.ac.uk/centres/csn/research-2/spike-sorting.
[21] Feldman, Daniel E. "The spike-timing dependence of plasticity." Neuron 75.4 (2012): 556-571.
[22] Gupta, Ankur, and Lyle N. Long. "Character recognition using spiking neural networks." Neural Networks, 2007. IJCNN 2007. International Joint Conference on. IEEE, 2007.
[23] MLA Lewicki, Michael S. "A review of methods for spike sorting: the detection and classification of neural action potentials." Network: Computation in Neural Systems 9.4 (1998): R53-R78.
[24] Karkare, Vaibhav, Sarah Gibson, and Dejan Marković. "A 75-µw, 16-channel neural spike-sorting processor with unsupervised clustering." IEEE Journal of Solid-State Circuits 48.9 (2013): 2230-2238.
[25] Karkare, Vaibhav, Sarah Gibson, and Dejan Markovic. "A 130-µW, 64-Channel Neural Spike-Sorting DSP Chip." IEEE journal of solid-state circuits 46.5 (2011): 1214-1222.
[26] Saeed, Maryam, Amir Ali Khan, and Awais Mehmood Kamboh. "Comparison of Classifier Architectures for Online Neural Spike Sorting." IEEE Transactions on Neural Systems and Rehabilitation Engineering 25.4 (2017): 334-344.
[27] Chen, Yi, Enyi Yao, and Arindam Basu. "A 128-channel extreme learning machine-based neural decoder for brain machine interfaces." IEEE transactions on biomedical circuits and systems 10.3 (2016): 679-692.
[28] Quiroga, Rodrigo Quian. "Spike sorting." Current Biology 22.2 (2012): R45-R46.
[29]Rey, Hernan Gonzalo, Carlos Pedreira, and Rodrigo Quian Quiroga. "Past, present and future of spike sorting techniques." Brain research bulletin 119 (2015): 106-117.
[30] Paraskevopoulou, Sivylla E., et al. "Hierarchical Adaptive Means (HAM) clustering for hardware-efficient, unsupervised and real-time spike sorting." Journal of neuroscience methods 235 (2014): 145-156.
[31] Nuntalid, Nuttapod, Kshitij Dhoble, and Nikola Kasabov. "EEG classification with BSA spike encoding algorithm and evolving probabilistic spiking neural network." International Conference on Neural Information Processing. Springer, Berlin, Heidelberg, 2011.
[32] Anand Kumar Mukhopadhyay, Indrajit Chakrabarti, and Mrigank Sharad, "Real-time digitized neural-spike storage scheme in multiple channels for biomedical applications", Asia-Pacific Signal and Information Processing Association Annual Summit and Conference (APSIPA ASC), 2017, pp. 1430-1435.
[33] Mukhopadhyay, Sudipta, and G. C. Ray. "A new interpretation of nonlinear energy operator and its efficacy in spike detection." IEEE Transactions on biomedical engineering 45.2 (1998): 180-187.